\documentclass[conference]{IEEEtran}
\IEEEoverridecommandlockouts
\usepackage{cite}
\usepackage{amsmath,amssymb,amsfonts}
\usepackage{graphicx}
\usepackage{textcomp}
\usepackage{xcolor}
\def\BibTeX{{\rm B\kern-.05em{\sc i\kern-.025em b}\kern-.08em
    T\kern-.1667em\lower.7ex\hbox{E}\kern-.125emX}}
\usepackage{amsmath,amsfonts,bm}

\def\eqref#1{equation~\ref{#1}}
\def\1{\bm{1}}

\DeclareMathAlphabet{\mathsfit}{\encodingdefault}{\sfdefault}{m}{sl}
\SetMathAlphabet{\mathsfit}{bold}{\encodingdefault}{\sfdefault}{bx}{n}

\usepackage{hyperref}
\usepackage{url}
\usepackage{cite}
\usepackage{amsmath,amssymb,amsfonts}
\usepackage{algorithm}
\usepackage{algorithmic}
\usepackage{textcomp}
\usepackage{booktabs}
\usepackage{multirow}
\usepackage{subfigure}

\begin{document}

\title{FedCM: A Real-time Contribution Measurement Method for Participants in Federated Learning\\
\thanks{This paper is supported in part by National Key Research and Development Program of China (No.2019YFB2102100), the Science and Technology Development Fund of Macau SAR (No. 0015/2019/AKP), the national Natural Science Foundation of China (Grant No. 61762033) and Hainan Provincial Natural Science Foundation of China (Grant No. 2019RC041 and 2019RC098).}
}

\author{\IEEEauthorblockN{1\textsuperscript{st} Bingjie Yan}
\IEEEauthorblockA{\textit{School of Compute Science and Cyberspace Security}\\
\textit{Hainan University}\\
Haikou, China \\
bj.yan@ieee.org}
\and
\IEEEauthorblockN{co-1\textsuperscript{st} Boyi Liu\thanks{Corresponding author: Boyi Liu, Email: by.liu@ieee.org}\thanks{Boyi Liu and Bingjie Yan contribute equally to this work.}}
\IEEEauthorblockA{\textit{State Key Laboratory of Internet of Things for Smart City} \\
\textit{University of Macao}\\
Macau SAR, China \\
by.liu@ieee.org}
\and
\IEEEauthorblockN{2\textsuperscript{nd} Yize Zhou}
\IEEEauthorblockA{\textit{School of Science} \\
\textit{Hainan University}\\
Haikou, China \\
Yz.zhou@ieee.org}

\and
\IEEEauthorblockN{3\textsuperscript{rd} Zhixuan Liang}
\IEEEauthorblockA{\textit{Department of Computing} \\
\textit{The Hong Kong Polytechnic University}\\
HongKong, China \\
zhixuan.liang@connect.polyu.hk}
\and
\IEEEauthorblockN{4\textsuperscript{th} Cheng-Zhong Xu}
\IEEEauthorblockA{\textit{Faculty of Science and Technology} \\
\textit{University of Macao}\\
Macau SAR, China \\
czxu@um.edu.mo}
}

\maketitle

\begin{abstract}
Federated Learning (FL) creates an ecosystem for multiple agents to collaborate on building models with data privacy consideration. The method for contribution measurement of each agent in the FL system is critical for fair credits allocation but few are proposed. In this paper, we develop a real-time contribution measurement method FedCM that is simple but powerful. The method defines the impact of each agent, comprehensively considers the current round and the previous round to obtain the contribution rate of each agent with attention aggregation. Moreover, FedCM updates contribution every round, which enable it to perform in real-time. Real-time is not considered by the existing approaches, but it is critical for FL systems to allocate computing power, communication resources, etc. Compared to the state-of-the-art method, the experimental results show that FedCM is more sensitive to data quantity and data quality under the premise of real-time. Furthermore, we developed federated learning open-source software based on FedCM. The software has been applied to identify COVID-19 based on medical images.
\end{abstract}

%\begin{IEEEkeywords}
%federated learning, contribution measurement, security
%\end{IEEEkeywords}

\section{Introduction}
Increasing developments of artificial intelligence make machine learning services become parts of our daily lives. Numerous Data are generated, collected, and accessed every day by smart terminals. Nevertheless, it is usually challenging to share them considering the privacy. For instance, it is impractical to use the language model trained by centralizing private data to predict the next word or even the entire reply because of data privacy\cite{ion2016metamaterial}. The emergence of federated learning (FL) breaks barriers among agents. FL enables terminal devices to use their local computing capabilities to coordinate model training while keeping data locally.
\begin{figure}[tp]
	\setlength{\abovecaptionskip}{0.cm}
	\begin{center}
		\includegraphics[width=3.0in,keepaspectratio]{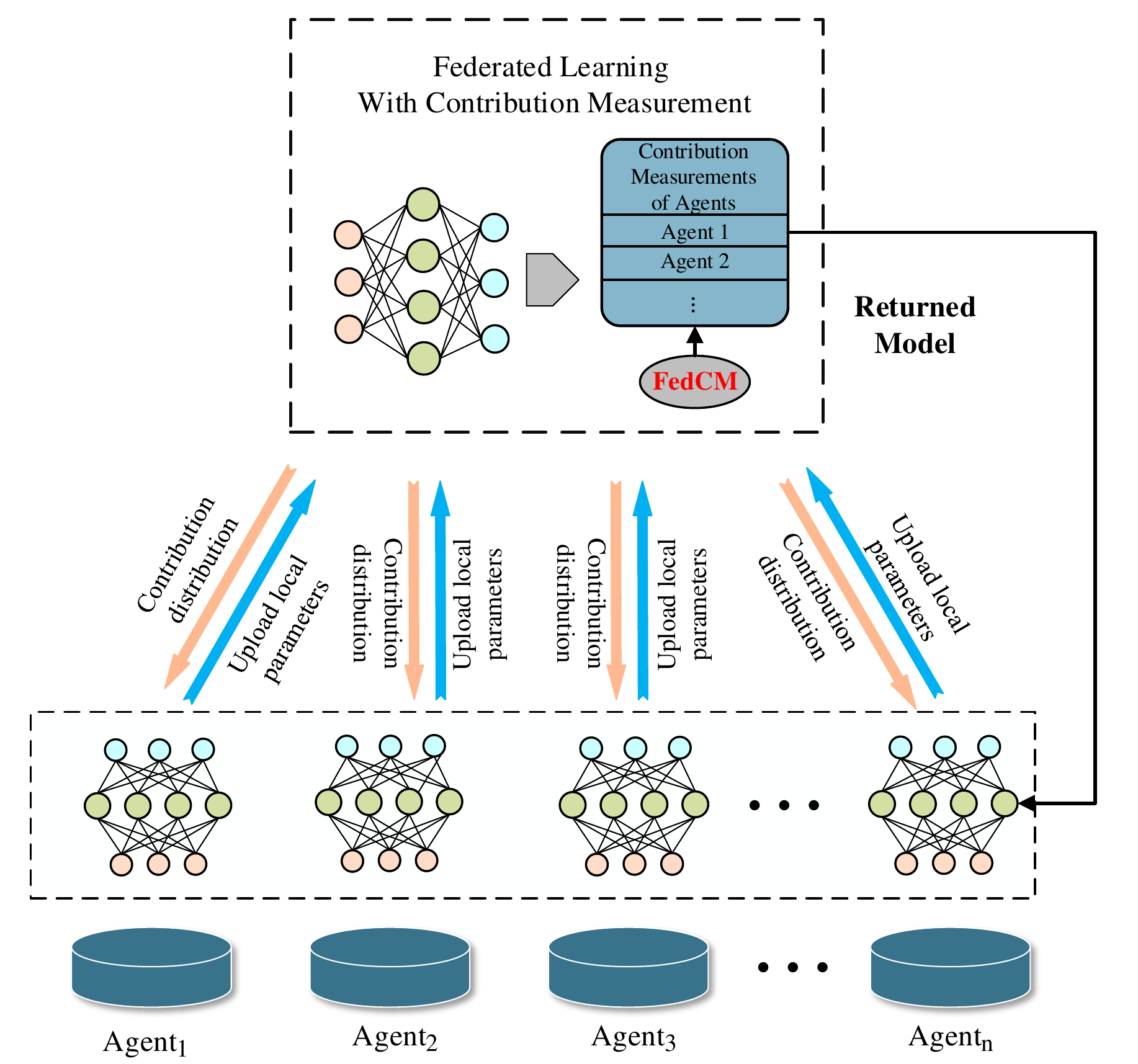}
	\end{center}
	\caption{Federated learning framework with FedCM.}
	\label{fl}
\end{figure}

FL system needs to encourage different agents to contribute their data and participate in the collaboration federation. Rewarding mechanism is crucial for the incentives current and potential participants of FL. Before that, FL needs a fair evaluation mechanism to give agents reasonable rewards. Each agent in FL should get corresponding rewards based on its contribution to the shared model, not the same reward. There are several proposed methods for contribution measurement. Such as \cite{wang2019measure} measured the contribution of each group features in vertical federated learning, and \cite{zhan2020learning} proposed an incentive mechanism to stimulate each agent to contribute better data in both quality and quantity. However, most of them consume considerable computing resources and calculate offline. Therefore, the existing methods are not capable of performing in real-time in FL systems.

The performance of real-time is critical to FL systems. Federated learning is a type of loosely distributed machine learning, which requires timely measurements of the contributions of the agents to allocate rewards such as computing resources, communication resources, etc. The existing computing-intensive and offline methods are difficult to deploy on FL systems. 

To address the issue, we develop a real-time contribution measurement method (FedCM) for participants in federated learning. The proposed method is simple but powerful. In the method, the attention mechanism of model aggregation is used to calculate the attention weight of all agents in FL. Each agent trains locally and uploads the parameters to the server. The server then adjusts to the centralized model and distributes it to the agents. ``Attention'' is assigned to each agent. The server then updates the parameters based on the ``attention'' of each agent. In FedCM, the method calculates the attention of each agent on the server, and get the attention of each agent on each layer of the centralized model. Then, the initial and final values of the server parameters are used to obtain the changed value of each parameter of the server. Finally, a mathematical formula is established to solve using the attention-based model\cite{hochreiter1997long}. The evaluator can compare the ideal results in the data transfer update to determine the contributions of all agents, and measure the contribution of each agent. 

FedCM is convenient to deploy in the FL system, and we have actually developed a Federated Learning software based on this. More meaningfully, we open source the software\footnote{FedMedical is available at \url{https://github.com/beiyuouo/paddle-fl-gui}} and it has been applied to medical image recognition of COVID-19.

Our contributions in the work are summarized as follows:
\begin{itemize}
	\item We propose a method to measure contributions of agents in the FL system, which consumes less computing resources and performs in real time. And we open sourced the method\footnote{Code is available at \url{https://github.com/beiyuouo/FedCM}}.
	\item The work conducts training for natural language processing  with federated learning. The method performs more sensitive to the quantity and quality of data of agents in FL system than the state-of-the-art method.
	\item An open sourced federated learning software is developed. The software has been applied to medical images recognition of COVID-19.
	%\item The method we propose is sensitive to data volume and data quality, and can be used for mutual comparison between agents.
	%\item In the training process, the contribution to each agent can be obtained in real time, with low computational complexity.
\end{itemize}

In the next chapters of this paper, we first briefly introduce Federated Learning. We then introduce the proposed method. The forth chapter presents experiments. The software "FedMedical" presents in chapter five. We conclude the paper with future work in the last chapter.

\section{Related Work}
In this section, we will introduce federated learning and its applications briefly. In addition, it also introduce the latest contribution measurement algorithm for participants in the federated learning system.
%This paper focus on the work of contribution measure mechanism in federated learning. Companies applying federated learning can apply this contribution incentive mechanism in the case of commission distribution based on agents contribution.

\subsection{Federated Learning}

Distribute the training data on each mobile device to maintain the localization of the data, instead of transmitting the data to the central server, updating the model locally, and uploading the update results to the server. While maintaining data localization and privacy, it can aggregate the data of each agent. The central server collects agent data and uses FedSGD, FedAVG\cite{brendan2016communication} and other algorithms to maintain the centralized model in combination with the different optimizer\cite{felbab2019optimization,ji2019learning}, and sends the updated model to each agent. During the transmission process, methods such as homomorphic encryption are used to protect the security of data transmission and maintain the continuous iterative update of the model. This method is federated learning.

At present, for different datasets, federated learning framework can be classified into horizontal federated learning, vertical federated learning, and federated transfer learning\cite{10.1145/3298981}. Horizontal federated learning is suitable for situations where the data provided by the agents has more of the same characteristics. In \cite{mcmahan2017communication}, Google proposed a solution to update the horizontal federated learning model in Android phones. vertical federated learning is suitable for situations where there is less feature overlap but more user id overlap. \cite{hardy2017private} proposed a vertical federated learning method for training a logistic model. \cite{chen2020fedhealth} proposed the FedHealth method, which uses federated learning to aggregate data and uses transfer learning to obtain a personalized model.

Under the framework of federated learning, there are many different directions of research. \cite{aono2017privacy} use the data transmitted by the homomorphic encryption agents and the central server for model training, which further strengthens the privacy of the agents' data. In \cite{lu2019blockchain, kim2019blockchained}, data verification is carried out in conjunction with the blockchain to prevent the gradient information from being maliciously tampered with. \cite{konevcny2016federated} research on reducing the consumption of communication resources in federated learning.

%The training data will be distributed on each mobile device, not all of them will be sent to the central server, and only the updated data on each device will be aggregated to the central server. After federated optimization, the central server returns to the global state of each device, and continues to accept the updated model parameters calculated by each agent in the new global state. This method is Federated Learning\cite{brendan2016communication}. Federated Learning \cite{konevcny2016federated} can solve the problem of unprotected large-scale private data and complete updating learning of devices without exchanging large amounts of data.

%This centralized training model approach provides privacy, security, regulation, and economic benefits\cite{zhao2018federated}. Federated Learning presents new statistical and system challenges when training models on distributed device networks\cite{smith2017federated}. Federated Learning, which relies on scattered data, brings many aspects of research: Fei Chen et al.\cite{chen2018federated} identified the combination of Federated Learning and Meta-learning as a major advance in Federated Learning. Konstantin Sozinov et al.\cite{sozinov2018human} have made some progress in applying Federated Learning to human activity identification.

\subsection{Application and Commercialization of Federated Learning}

Since federated learning was proposed, federated learning has been successfully applied to more and more scenarios. When considering data privacy issues, many companies will choose to use federated learning to protect data privacy to achieve cooperation. Such WeBank has successfully used federated learning in bank federations for credit evaluation and other financial aspect. In \cite{ren2019federated}, federated learning is applied to dynamic IoT system.\cite{lu2019blockchain, kim2019blockchained} use blockchain for data verification as an alternative to the commercialization of federated learning. \cite{bao2019flchain} achieved model commercialization by providing a healthy marketplace for collaborative-training models.

In the commercialization process of federated learning, the method of using this decentralized training model provides privacy, security, supervision and economic benefits, but also brings fairness challenges. Therefore, a fair contribution measurement method is essential for FL system.

\subsection{Contribution Measurement and Incentive Mechanism in Federated Learning}

\cite{wang2019measure} proposed grouped instance influence to measure contribution of agents in horizontal federated learning and grouped feature importance in vertical federated learning. This approach equitably and accurately realized the measurement of contributions of all agents but it cannot take into account the quantity of data.

\cite{geyer2017differentially} proposed a federated learning optimization algorithm based on differential privacy protection of the agents that can hide the contribution of the agents. In order to balance the contribution of the soft training model and ensure the collaborative convergence, \cite{xu2019elfish} proposed the corresponding parameter aggregation scheme. \cite{kang2019incentive} used the weighted subjective logic model to design a reputation-based worker selection plan to achieve reliable federated learning. In \cite{zhan2020learning}, the authors designed an incentive mechanism based on deep reinforcement learning to determine the optimal pricing strategy for parameter servers and the optimal training strategy for agents.

Our work focus on real-time contribution measurement mechanism for each agent in vertical federated learning. Current researches have problems such as poor real-time performance and high resource consumption by contribution measurement methods. The method proposed in this paper can obtain real-time contribution measurement to each agent in the process of federated learning. At the same time, it verifies that the method proposed in this paper has high sensitivity on data quantity and data quality.

\section{Methodology}

In this section, we will introduce the basic framework of federated learning and how parameters are updated. Based on this, we then introduce the proposed method in detail.

\subsection{The General Federated Learning Approach}
Federated learning is a distributed learning method in which the cloud(server) maintains a centralized model and distributes it to individual agents. Rather than uploading agents' data and training models in the cloud, FL lets agents train a separate model with their own and upload those black-box model parameters to the server and fuse the models. The server sets up a fraction $C$ to select the agents proportionally for the server's centralized model update. The server updates the centralized model, and distributes the model to the agents. This not only avoids the transmission and storage of user’s sensitive personal data, but also utilizes the computational power on the agents and reduce the computation pressure from their central server \cite{wang2019measure}. 
Federated averaging \cite{mcmahan2017communication} algorithm is the basic method of federated learning, which can be expressed as Algorithm \ref{[Algorithm 1]}.

%Federated learning is a distributed learning method in which the server maintains an central model and distributes it to individual agents. To solve privacy problems, the server does not need to obtain the data of the agents, so the computing power of the agents is used to learn at the agents local environment.

%The server sets up a fraction $C$, to select the agents proportionally for the server's central model update. Then the selected agents' updated model parameters are uploaded to the server for parameters updating of the central model. It is then distributed to the each agent to improve the model of the agents.

%In this way, it continue to improve the central model of the server and the local model of the agents. Under the premise of ensuring the correctness and privacy of the agents, using this method can make full use of the computing power of the agents and a large amount of private data for learning, and at the same time maintain an excellent global model.

\begin{algorithm}
	\caption{Federated Averaging Algorithm}
	\label{[Algorithm 1]}
	\begin{algorithmic}[1]
		\STATE \emph{\textbf{Server model update}}
		\STATE $K$ is the total number of agents, $w_t$ is the parameters of current central server model, $w_t^k$ is the parameters of current agents’ model with index $k$.
		\STATE \hspace*{0.02in}{\bf{Input:}} server parameters $w_{t}$ at $t$, agent parameters $w_{t+1}^{k}$ at $t+1$
		\STATE \hspace*{0.02in}{\bf{Output:}} server parameter $w_{t+1}$ at $t+1$
		\STATE Initialize $w_0$
		\FOR{each round $t=1,2,\cdots$ }
		\STATE $m\leftarrow max\{C\cdot K, 1\}$
		\STATE $s_t\leftarrow$ random set of $m$ agents from all $K$ agents 
		\FOR{each $k\in S_{t}$}
		\STATE $w_{t+1}^k\leftarrow$ ClientUpdate($k$, $w_{t}^k$) //agents update model on local device
		\ENDFOR
		\STATE $w_{t+1}\leftarrow$ ServerOptimization($w_t$,$w_{t+1}^k$)
		\ENDFOR
	\end{algorithmic}

	% \rule[-10pt]{8.7cm}{0.05em}
	\begin{algorithmic}[1]
		\STATE \emph{\textbf{Local model update}}
		\STATE $K$ is the total number of agents, $B$ is the local mini-batch size, $E$ is the number of local epochs, $S$ is a set of all agents, $w_t$ is the parameters of current central server model, $w_t^k$ is the parameters of current agents’ model with index $k$, and $\eta$ is the learning rate.
		\STATE \hspace*{0.02in}{\bf{Input:}} index of client $k$, user private data $x_k$
		\STATE \hspace*{0.02in}{\bf{Output:}} client parameters $w_{t+1}^k$
		\STATE $B \leftarrow$ Split user data into local mini-batch size $B$
		\FOR{each local epoch $e$ from $1$ to $E$}
		\FOR{batch $b \in B$}
		\STATE $w\leftarrow w-\eta \bigtriangledown L(w;b)$
		\ENDFOR
		\ENDFOR
		\RETURN $w$ to server
	\end{algorithmic}
\end{algorithm}  

\subsection{Attention Aggregation in FedCM}
%In this section, we will introduce a FedAtt \cite{ji2019learning}, which is outperformed to the common FedAvg \cite{} and FedSGD \cite{} algorithm.

FedCM runs on the central server of federated learning system, and performs after receiving the parameters of each agent every round or every several rounds. It utilizes attention mechanism. The server of Federated Learning with FedCM uses \eqref{[Formula 1]} to calculate the value of attention $\alpha$ of each layer parameter that should be allocated to the agent, and multiplies each layer parameter by the corresponding attention to update the centralized model. The server aggregates the update parameters uploaded by the agents, and finally completes the update of the local model through the downlink communication between the server and the agent. The process of attention aggregation in FedCM can be summarized as the following formulas and the attention aggregation function in Algorithm \ref{[Algorithm 2]}.

\begin{equation}
\label{[Formula 1]}
\alpha_{k}^{l}=\text{softmax}(s_{k}^{l})=\frac{e^{s_{k}^{l}}}{\sum_{i}e^{ s_{i}^{l}}}
\end{equation}

The $s_{k}^{l}$ is the norm difference from the central model. We use $w^{l}$ to represent layer $l$ parameters of the server and $w_{k}^l$ for $l$th layer parameters of the agent model $k$. So the definition of $s_{k}^{l}$ is as follows 
\begin{equation}
s_{k}^{l}=||w^{l}-w_{k}^{l}||_{p}
\end{equation}

%So for $m$ agents that are selected to update the central model, the method of aggregation becomes
%\begin{equation}
%w_{t+1}\leftarrow w_{t}-\epsilon \sum_{k=1}^{m}\bigtriangledown (w_{t}^{k})=w_{t}-\epsilon \sum_{k=1}^{m}\alpha_{k}(w_{t}-w_{t}^{k})
%\end{equation}

To protect the agents’ data privacy, you can add the randomized mechanism before the agent passes parameters to the server. Randomly generate a random vector that obeys the standard distribution $\mathcal {N}(0,\sigma^2)$, multiply the corresponding weight $\beta$, the results of the final update parameters as shown in \eqref{[Formula 2]}.
\begin{equation}
\label{[Formula 2]}
w_{t+1}\leftarrow w_{t}-\epsilon \sum_{k=1}^{m}\alpha_{k}(w_{t}-w_{t}^{k}+\beta \mathcal {N}(0,\sigma^2 ))
\end{equation}

Attention calculation is the first step of FedCM. The method performs contributions quantification based on the parameters obtained from attention aggregation.
%The implementation process of the whole algorithm is shown as Algorithm %\ref{[Algorithm 2]}
%
%\begin{algorithm}
%	\caption{Attentive Federated Optimization}
%	\label{[Algorithm 2]}
%	\begin{algorithmic}[1]
%		\STATE $l$ is the ordinal of neural network layers; $\epsilon$ is the stepsize of server optimization
%		\STATE \hspace*{0.02in}{\bf Input:} server parameters $w_t$ at $t$, agents parameters $w_{t+1}^1, w_{t+1}^2,\cdots, w_{t+1}^{m}$ at $t+1$
%		\STATE \hspace*{0.02in}{\bf Output:} server parameters $w_{t+1}$ at $t+1$
%		\STATE Initialize attention $\alpha =\{\alpha_{1}, \alpha_{2},\cdots, \alpha_{m}\}$
%		\FOR{each layer $l$ in model}
%		\FOR{each agents $k \in S_{t}$ from $1$ to $m$}
%		\STATE $s_{k}^{l}=||w^{l}-w_{k}^{l}||_{p}$
%		\ENDFOR
%		\STATE $\alpha_{k}^{l}=softmax(s_{k}^{l})=\frac{e^{s_{k}^{l}}}{\sum_{i}e^{s_{i}^{l}}}$
%		\ENDFOR
%		\STATE $\alpha_{k}=\{\alpha_{k}^{1},\alpha_{k}^{2},\cdots,\alpha_{k}^{l}\}$
%		\STATE $w_{t+1}\leftarrow w_{t}-\epsilon \sum_{k=1}^{m}\alpha_{k}(w_{t}-w_{t}^{k}+\beta \mathcal {N}(0,\sigma^2 ))$
%	\end{algorithmic}
%\end{algorithm}  

\subsection{Contributions Quantification in FedCM}
This section introduces the procedure of contributions quantification in FedCM. Before that, we make the following assumptions: Each agent does not tamper with the updated gradient itself during the previous process of transmitting between the server and the agents. 

FedCM calculate each agent contribution to the system with each layer of agents' parameter attention. When the server has obtained the attention of agents $K$, it can calculate the impact of the agents $K$ on the parameters of the centralized model in the $T$-th step of update. We define $imp_ {t}^{k}$ as follows:

\begin{equation}
imp_{t}^{k}=\epsilon \alpha_{k}[\frac{\log{(w_{t}-w_{t}^{k})+1}}{\log{(w_{t+1}-w_{t})+1}}+\beta \mathcal {N}(0,\sigma^2 )]+\gamma\cdot imp_{t-1}^{k}
\end{equation}

where $\gamma \in (0,1)$ is forgetting coefficient, because large variations in the early stage of the centralized model, the model is not stable, and the impact of the previous endpoint should be reduced after multiple iterations. If the agent $k$ is not in the selected $M$ agents this time, we think that the impact of the agent $t$ round is $imp_{t}^{k}=imp_{t-1}^{k}$, that is, only the number of rounds that the agent participates in the update is calculated. Note that the server's attention to the agents are the attention of each layer, so when calculating the impact it also calculates the impact of each layer by reweighting and averaging.

FedCM normalizes agents' attention in a limited range with the MinMaxScaler and Softmax, to obtain the contribution of each agent. We use $con_{t}^{k}$ to represent the contribution of agent $k$ at the $t$ step.

\begin{algorithm}
	\caption{FedCM-drived fedearted learning}
	\label{[Algorithm 2]}
	\begin{algorithmic}[1]
	    %\COMMENT{$l$ is the ordinal of neural layers; $\epsilon$ is the stepsize of server optimization}
		\STATE \emph{\textbf{Attention Calculation}}
		\STATE $l$ is the ordinal of neural layers; $\epsilon$ is the stepsize of server optimization
        \STATE \hspace*{0.02in}{\bf Input:} server parameters $w_t$ at $t$, agents parameters $w_{t+1}^1, w_{t+1}^2,\cdots, w_{t+1}^{m}$ at $t+1$
        \STATE \hspace*{0.02in}{\bf Output:} server parameters $w_{t+1}$ at $t+1$
        %\STATE \hspace*{0.02in}{\bf procedure} ATTENTIVE OPTIMIZATION
        \STATE Initialize attention $\alpha =\{\alpha_{1}, \alpha_{2},\cdots, \alpha_{m}\}$
        \FOR{each layer $l$ in model do}
        \FOR{each agents $k \in S_{t}$ from $1$ to $m$ do }
            \STATE $S_{k}^{l}=||w^{l}-w_{k}^{l}||_{p}$
        \ENDFOR
        \STATE $\alpha_{k}^{l}=softmax(s_{k}^{l})=\frac{e^{s_{k}^{l}}}{\sum_{i}e^{s_{i}^{l}}}$
    \ENDFOR
    \RETURN $\alpha_{k}=\{\alpha_{k}^{1},\alpha_{k}^{2},\cdots,\alpha_{k}^{l}\}$
    %\STATE $w_{t+1}\leftarrow w_{t}-\epsilon\sum_{k=1}^{m}\alpha_{k}(w_{t}-w_{t}^{k}+\sbeta \mathcal {N}(0,\sigma^2 ))$
	\end{algorithmic}
	
	\begin{algorithmic}[1]
	%\Procedure{Contributions Quantification}
	    \STATE \emph{\textbf{Contributions Quantification}}
	    \STATE \hspace*{0.02in}{\bf Input:} the ordinal of neural network layers $l$; the stepsize of server optimization $\epsilon$, server parameters $w_{t}$ at time $t+1$, agents parameters $w_{t+1}^1, w_{t+1}^2,\cdots, w_{t+1}^{m}$ at time $t+1$, forgetting coefficient $\gamma$,  selected agents set $S_t$.
		\STATE \hspace*{0.02in}{\bf Output:} Agents contributions $con_{t}$ at $t$
		\STATE \emph{Attention Calculation($w_t, w_{t+1}^{k}$)}
		\FOR{each agents $k$ from $1$ to $K$}
		\IF{$k\in S_{t}$}
		\STATE $imp_{t}^{k}=\epsilon \alpha_{k}[\frac{\log{(w_{t}-w_{t}^{k})+1}}{\log{(w_{t+1}-w_{t})+1}}+\beta \mathcal {N}(0,\sigma^2 )]+\gamma\cdot imp_{t-1}^{k}$
		\ELSE
		\STATE $imp_{t}^{k}=imp_{t-1}^{k}$
		\ENDIF
		\ENDFOR
		\RETURN $con_{t}=\text{Softmax}(\text{MinMaxScaler}(imp_{t}))$
	\end{algorithmic}
\end{algorithm}

\section{EXPERIMENT}
In this section, we will introduce the verification experiments performed for our proposed agents' contribution measurement method.

\subsection{Experiment for Image Classification}

\subsubsection{Experimental Environment and Models}
The system used in experiments is Ubuntu 18.04 LTS, the backend used is the GPU version of Pytorch, with the NVIDIA GTX1660Ti GPU acceleration. We use ResNet to perform experiments for images classification with datasets of MNIST\footnote{MNIST is available at \url{http://yann.lecun.com/exdb/mnist}} and FashionMNIST\footnote{FashionMNIST is available at \url{https://github.com/zalandoresearch/fashion-mnist}} to verify the effectiveness of FedCM and its sensitivity to special variables. At the same time, we believe that the data of each agent is independent and identically distributed. The parameters are listed in Tab. \ref{tab1}.

\begin{table}[htbp]
	\caption{Parameters}
	\label{tab1}
	\begin{center}
		\begin{tabular}{ccc}
			\toprule
			\textbf{Name}& \textbf{Represent letter}& \textbf{Value} \\
			\midrule % \hline \\
			Number of agents & $K$ & $10$ \\ %\hline
			Server training rounds & round & $10$ \\ %\hline
			The number of local epoch & epoch & $1$ \\ %\hline
			The fraction of agents & $C$ & $1$ \\ %\hline
			% Learning rate of agents & $\eta$ & $0.01$ \\ %\hline
			% Learning rate of server & $\eta'$ & $0.025$ \\ %\hline
			Batch size & $B$ & $128$ \\ %\hline
			Step size & $\epsilon$ & $1.2$ \\ %\hline
			Differential privacy & $\beta$ & $0.001$ \\ %\hline
			Forgetting coefficient & $\gamma$ & $0.7$ \\ %\hline
			\bottomrule
		\end{tabular}
	\end{center}
\end{table}

\begin{table*}[h!t]
\caption{Experiment Result of Image Classification}
\resizebox{2.0\columnwidth}{!}{
    \begin{tabular}{ccccc}
    \toprule
    \multicolumn{1}{p{2.875em}}{Model} & \multicolumn{1}{p{3.94em}}{Option} & \multicolumn{1}{p{5.75em}}{Dataset} & Method & Contributions \\
    \midrule
    \multicolumn{1}{c}{\multirow{12}[24]{*}{ResNet}} & \multicolumn{1}{c}{\multirow{4}[8]{*}{Normal}} & \multicolumn{1}{c}{\multirow{2}[4]{*}{MNIST}} & FedCM & \textbf{0.059;0.161;0.087;0.070;0.074;0.077;0.121;0.144;0.134;0.073} \\
\cmidrule{4-5}          &       &       & Shapley & 0.100;0.100;0.104;0.099;0.099;0.103;0.099;0.098;0.102;0.096 \\
\cmidrule{3-5}          &       & \multicolumn{1}{c}{\multirow{2}[4]{*}{FashionMNIST}} & FedCM & \textbf{0.091;0.097;0.126;0.082;0.054;0.143;0.074;0.107;0.146;0.081} \\
\cmidrule{4-5}          &       &       & Shapley & 0.099;0.101;0.099;0.099;0.099;0.100;0.099;0.101;0.100;0.102 \\
\cmidrule{2-5}          & \multicolumn{1}{c}{\multirow{4}[8]{*}{Random}} & \multicolumn{1}{c}{\multirow{2}[4]{*}{MNIST}} & FedCM & \textbf{0.112;0.111;0.112;0.112;0.112;0.111;0.111;0.111;0.067;0.041} \\
\cmidrule{4-5}          &       &       & Shapley & 0.110;0.110;0.110;0.110;0.109;0.109;0.109;0.109;0.062;0.062 \\
\cmidrule{3-5}          &       & \multicolumn{1}{c}{\multirow{2}[4]{*}{FashionMNIST}} & FedCM & \textbf{0.113;0.113;0.113;0.113;0.113;0.113;0.113;0.113;0.042;0.052} \\
\cmidrule{4-5}          &       &       & Shapley & 0.106;0.105;0.106;0.106;0.105;0.105;0.106;0.106;0.078;0.077 \\
\cmidrule{2-5}          & \multicolumn{1}{c}{\multirow{4}[8]{*}{Mislabeled}} & \multicolumn{1}{c}{\multirow{2}[4]{*}{MNIST}} & FedCM & \textbf{0.114;0.114;0.114;0.114;0.114;0.114;0.114;0.114;0.046;0.042} \\
\cmidrule{4-5}          &       &       & Shapley & 0.110;0.109;0.110;0.109;0.109;0.110;0.109;0.109;0.063;0.062 \\
\cmidrule{3-5}          &       & \multicolumn{1}{c}{\multirow{2}[4]{*}{FashionMNIST}} & FedCM & \textbf{0.110;0.109;0.110;0.109;0.109;0.110;0.109;0.109;0.063;0.062} \\
\cmidrule{4-5}          &       &       & Shapley & 0.106;0.106;0.105;0.105;0.106;0.105;0.105;0.106;0.077;0.078 \\
    \bottomrule
    \end{tabular}}
\end{table*}

\begin{figure}[htb]
	%\begin{minipage}[c]{0.5\textwidth}
	\begin{center}
		\subfigure[]{\includegraphics[width=1.6in, keepaspectratio]{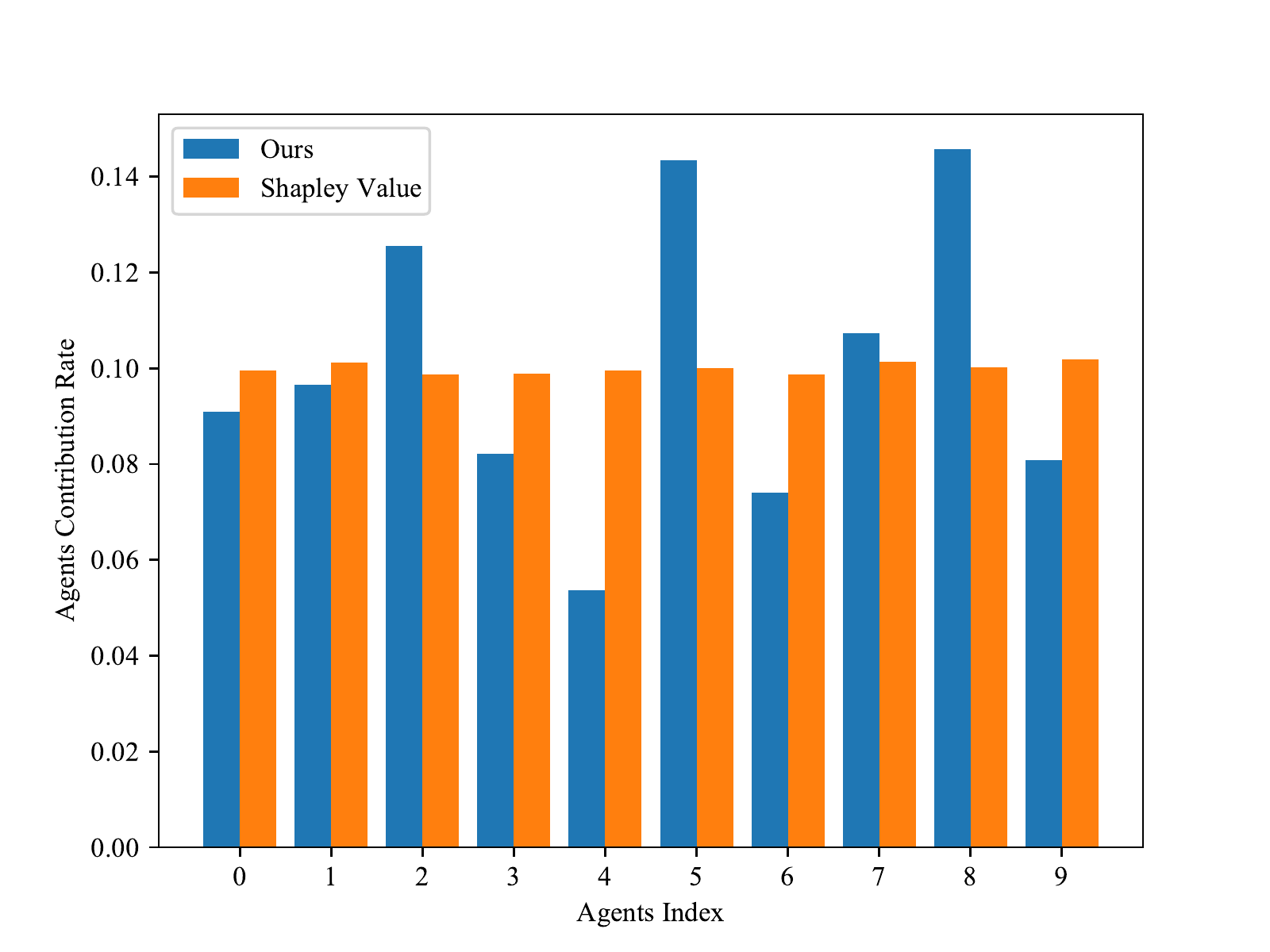}}
		\subfigure[]{\includegraphics[width=1.6in, keepaspectratio]{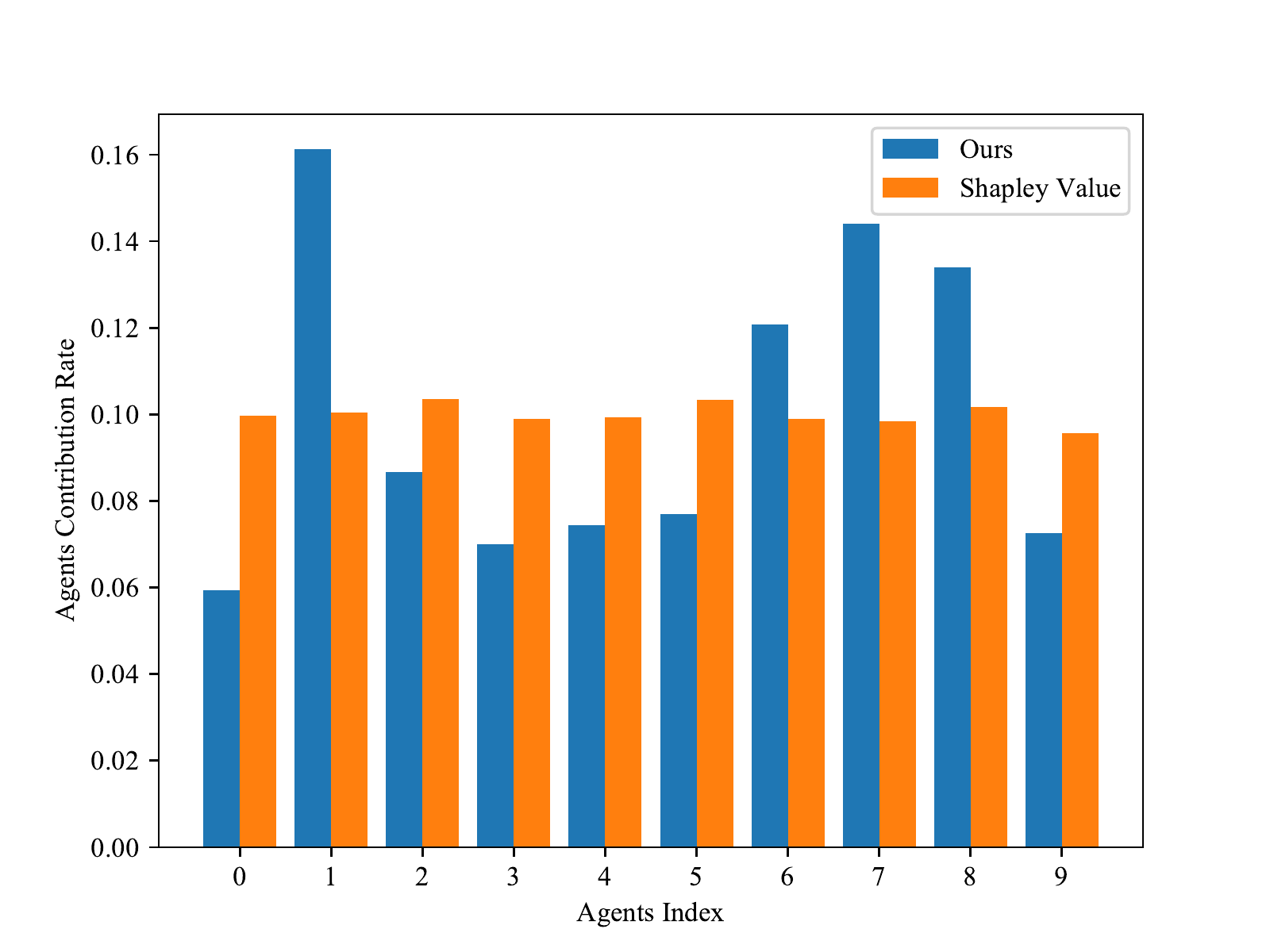}}
	\end{center}
	\caption{Experimental results of normal measurement on FashionMNIST(a) and MNIST(b).}
	\label{resnet-normal}
	%\end{minipage}
\end{figure}
\begin{figure}[htb]
    %\begin{minipage}[c]{0.5\textwidth}
	\begin{center}
		\subfigure[]{\includegraphics[width=1.6in, keepaspectratio]{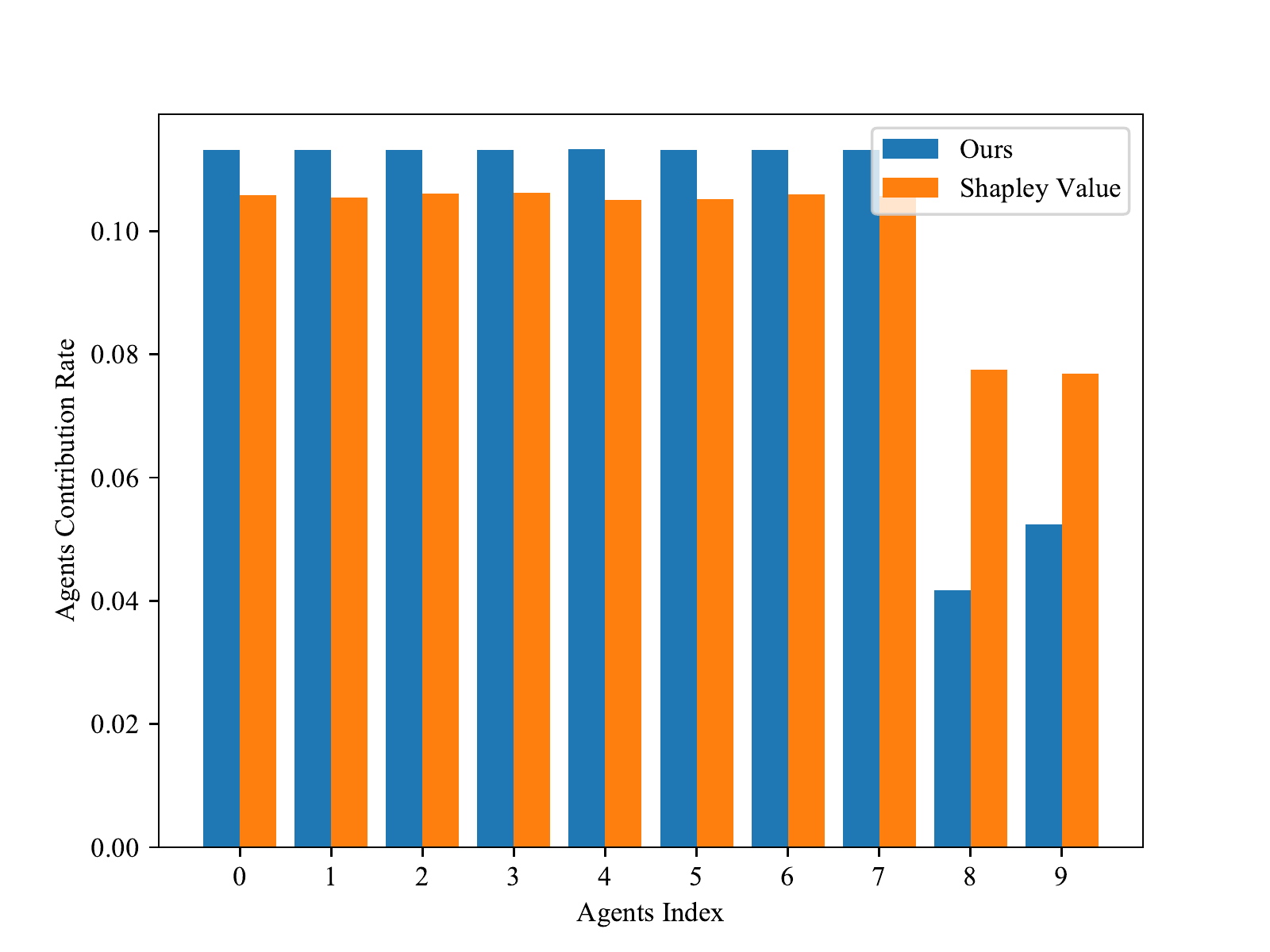}}
		\subfigure[]{\includegraphics[width=1.6in, keepaspectratio]{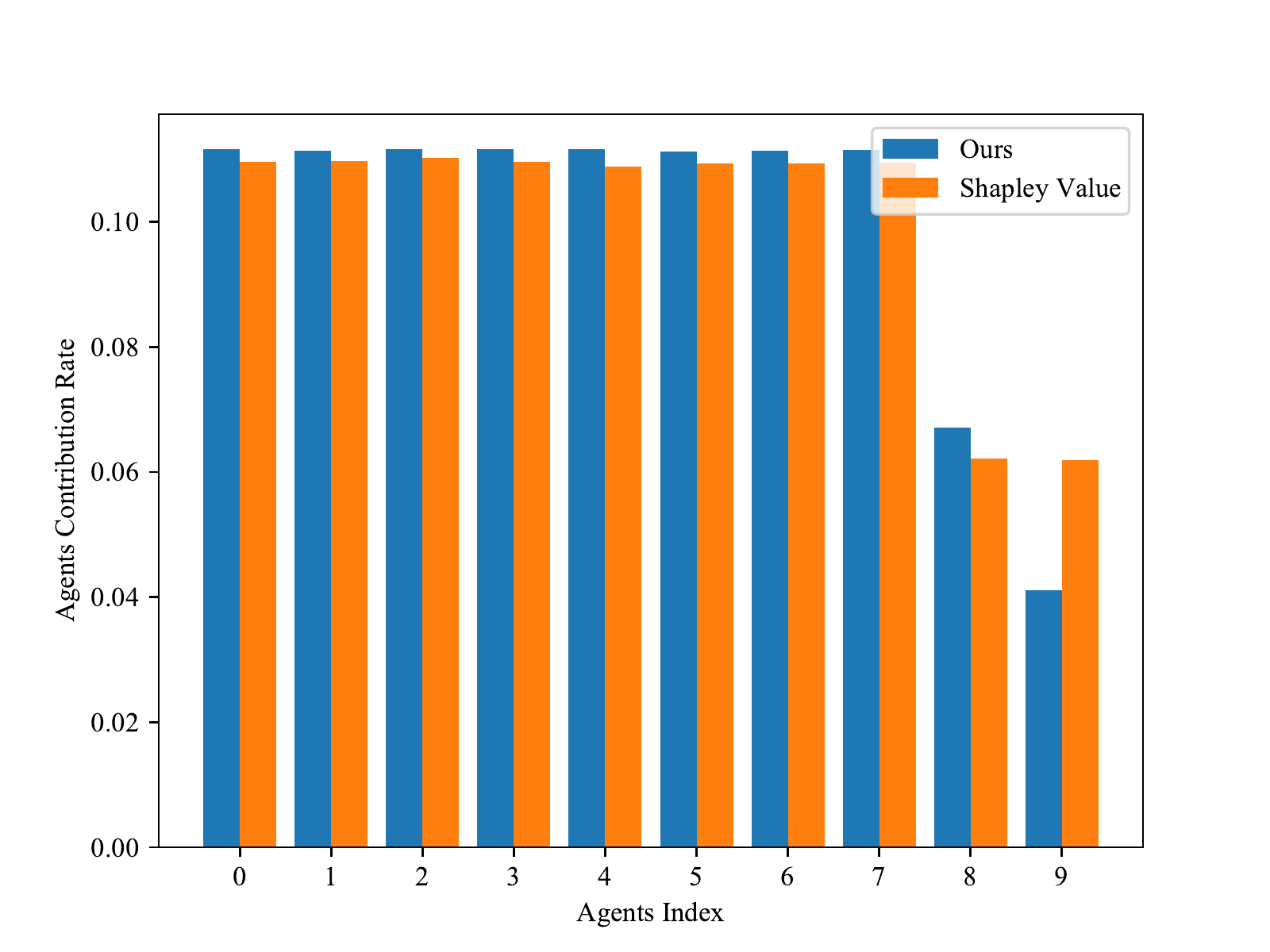}}
	\end{center}
	\caption{Experimental results of random noise case on FashionMNIST(a) and MNIST(b).}
	\label{resnet-random}
	%\end{minipage}
\end{figure}
\begin{figure}[htb]
	%\begin{minipage}[c]{0.5\textwidth}
	\begin{center}
		\subfigure[]{\includegraphics[width=1.6in, keepaspectratio]{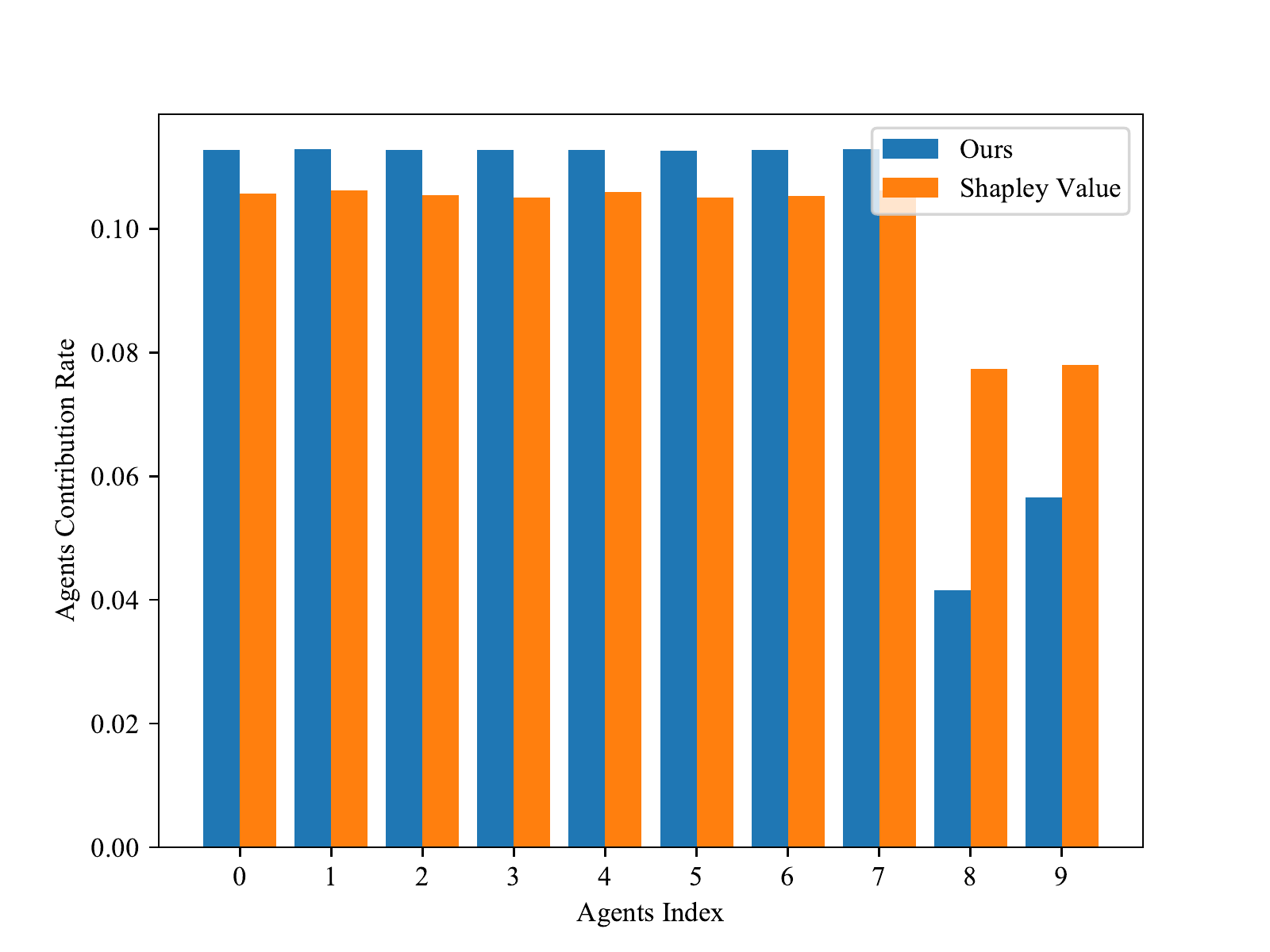}}
		\subfigure[]{\includegraphics[width=1.6in, keepaspectratio]{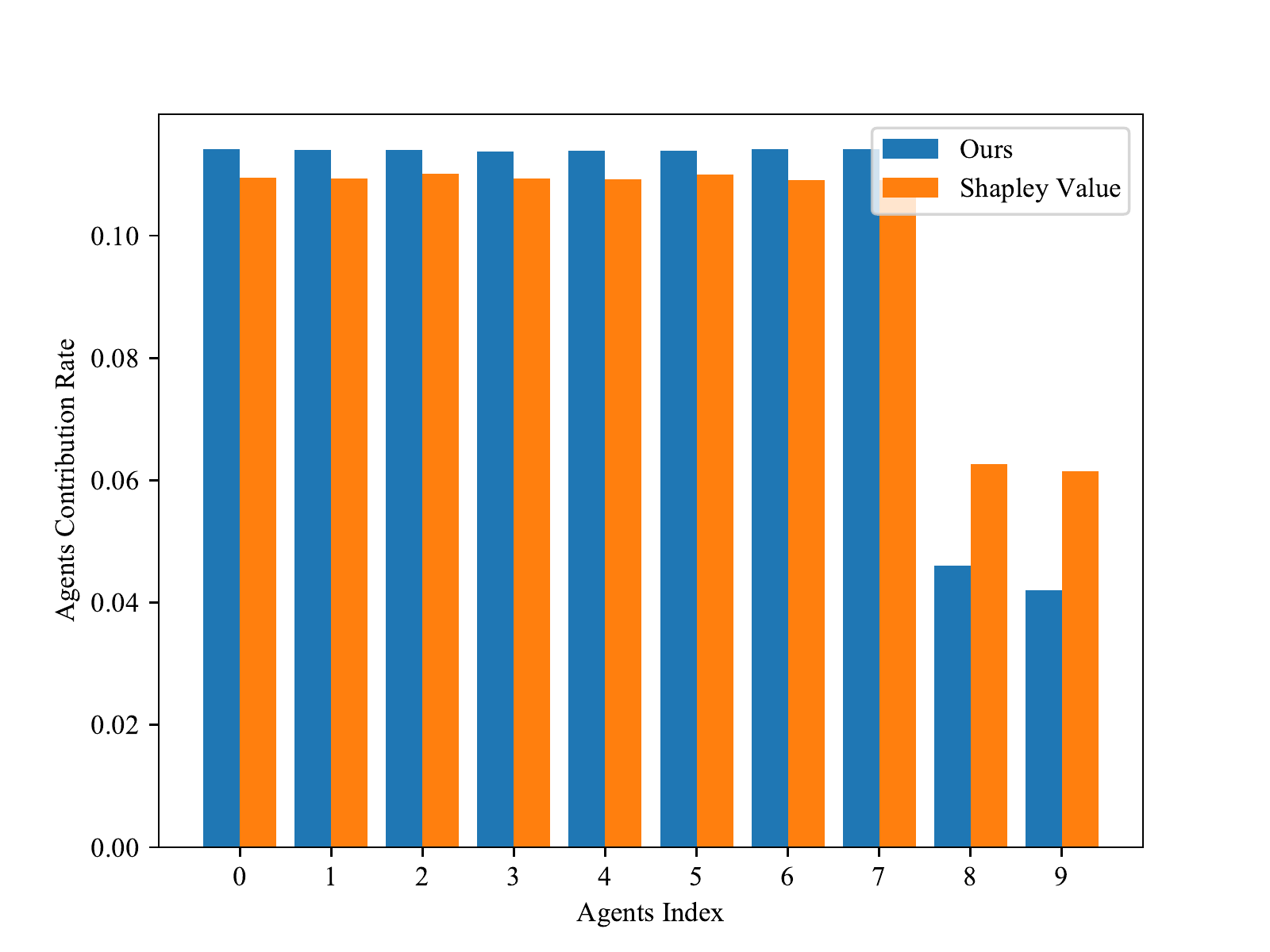}}
	\end{center}
	\caption{Experimental results of mislabeled case on FashionMNIST(a) and MNIST(b).}
	\label{resnet-mislabel}
	%\end{minipage}
\end{figure}

\subsubsection{Experimental Result}
We assumed three scenarios in image classification federation learning to verify the effectiveness of our method and sensitivity to data. And we compared it with the results of the Shapley Value evaluation. Shapley Value is originated from coalitional game theory and has proven theoretical properties. And the state-of-the-art work \cite{wang2019measure} proposed a FL measurement method based on Shaply Value. It provides a way to measure the impact and contribution of various agents. The definition of Shapley Value is:

\begin{equation}
\varphi_{i}(x)=\sum_{Q\subseteq S\setminus \{i\}} \frac{|Q|!-(|S|-|Q|-1)!}{|S|!}(\Delta_{Q\cup \{i\}}(x)-\Delta_{Q}(x))
\end{equation}

$S$ is the set of all agents, $Q\subset S={1,2,\cdots, n}$ is a subset of the agent set $S$, $i$ is the index of the agents, $|\cdot|$ represents the size of the set, $\Delta_{Q}(x)=imp_{Q}$ denotes the impact of agent set $Q$.
We use the following estimation method to get the $\varphi_{i}(x)$ of each agent because the complexity of directly calculating Shapley value is too high.

\begin{equation}
\varphi_{i}(x)=\frac{1}{M}\sum_{m=1}^{M}(\Delta_{Q^{m}\cup \{i\}}(x)-\Delta_{Q^{m}}(x))
\end{equation}

where $M$ is the number of iterations. $\Delta_{Q^{m}}(x)$ denotes the impact of random set $Q^{m}$. Finally, all the data obtained by Shapley Value is subjected to the same regularization processing as ours.

\textbf{\emph{Note that even if use estimation method, the contribution of Shapley Value obtained must be recalculated after removing some agents, and FedCM does not require recalculation and can get real-time contributions after each round of centralized model aggregation. This mechanism enables FedCM perform in real-time during the federated learning process.}}\\
\textbf{-- Experimental results of the normal measurement}\\
We did not do any special treatment to any agents and randomly divided the dataset into each agent. The result is shown in Fig. \ref{resnet-normal}. It can be seen that the deviation of each agent is not large, only individual agents are too high or too low. And the result of Shapley Value is more even.\\
\textbf{-- Experimental results of the random noise case}

\begin{figure}[htb]
    \begin{minipage}[c]{0.5\textwidth}
	\begin{center}
		\includegraphics[width=1.0in, keepaspectratio]{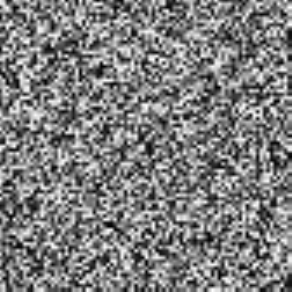}
	\end{center}
	\caption{Random noise data belonging to special agents.}
	\label{random-noise}
    \end{minipage}
\end{figure}

In this experiment, we modified the datasets of the last two agents' data to random noise, such as Fig. \ref{random-noise}. These should be regarded as dirty data by the model, so as to get a smaller return. The results are shown in Fig. \ref{resnet-random}. Both the method proposed in this paper and the Shapley Value method can identify these bad agents and give them a small contribution. The gap between normal agents and special agents is more obvious in our result.\\
\textbf{-- Experimental results of the mislabeled case}

In image classification scenarios, wrong labels may seriously affect the quality of the data. In this experiment, we randomly change the label in the special agents' data and the results are shown in Fig. \ref{resnet-mislabel}. The results are similar to the results of the previous experiment, and a better result than Shapley Value can be obtained, indicating that our method is more sensitive to such poor quality data.

\subsection{Experiment for Language Processing}
\subsubsection{Experimental Environment and Models}

The system environment is the same as above. But we add agents number to 20 and perform verification experiments for the language processing model on public language dataset of Penn Treebank\footnote{Penn Treebank is available at  \url{https://github.com/wojzaremba/lstm/tree/master/data}}
%  and WikiText-2\footnote{WikiText-2 is available at https://s3.amazonaws.com/research.metamind.io/wikitext/wikitext-2-v1.zip} 

In natural language processing, the RNN model is often used for processing. We used a smaller GRU-based agent language model. First, take texts as input, and convert words into word vectors according to a pre-built dictionary. The converted word vector is then used as an input to the RNN model, and the prediction result is finally output.

We use testing perplexity as an indicator of the evaluation model. In information theory, perplexity is a measurement of how well a probability distribution or probability model predicts a sample. 
The perplexity of a discrete probability distribution is defined as:
\begin{equation}
ppl(x)=2^{H(p)}=2^{-\sum_{x}p(x)log_{2}p(x)}
\end{equation}

In the above equation, $H(p)$ is the expected probability distribution. If the prediction result of our model is $m(x)$, then the perplexity of the language model is defined as:
\begin{equation}
ppl(x)=2^{H(p, m)}=2^{-\sum_{x}p(x)log_{2}m(x)}
\end{equation}

Obviously, $H(p)\leqslant H(p,m)$. Therefore, the smaller the perplexity, the more representative the probability distribution can be to better predict the sample distribution.

\subsubsection{Experiment Result}
\begin{table*}[h!t]
\caption{Experiment Result of Image Classification}
\resizebox{2.0\columnwidth}{!}{
    \begin{tabular}{ccccc}
    \toprule
    \multicolumn{1}{p{2.625em}}{Model} & \multicolumn{1}{p{3.19em}}{Option} & \multicolumn{1}{p{3em}}{Dataset} & Method & Contributions \\
    \midrule
    \multicolumn{1}{c}{\multirow{8}[16]{*}{RNN}} & \multicolumn{1}{c}{\multirow{2}[4]{*}{normal}} & \multirow{2}[4]{*}{ptb} & Ours  & \multicolumn{1}{c}{\textbf{0.060;0.028;0.032;0.046;0.032;0.043;0.068;0.071;0.075;0.051;0.035;0.060;0.044;0.036;0.071;0.039;0.039;0.071;0.046;0.053}} \\
\cmidrule{4-5}          &       &       & Shapley & 0.041;0.059;0.052;0.044;0.038;0.052;0.066;0.062;0.052;0.047;0.030;0.051;0.041;0.050;0.045;0.042;0.083;0.040;0.054;0.052 \\
\cmidrule{2-5}          & \multicolumn{1}{c}{\multirow{2}[4]{*}{less}} & \multicolumn{1}{c}{\multirow{2}[4]{*}{ptb}} & Ours  & \multicolumn{1}{c}{\textbf{0.057;0.050;0.048;0.054;0.051;0.053;0.059;0.063;0.068;0.061;0.054;0.054;0.059;0.052;0.062;0.055;0.025;0.025;0.025;0.025}} \\
\cmidrule{4-5}          &       &       & Shapley & 0.044;0.057;0.049;0.059;0.086;0.045;0.057;0.042;0.047;0.041;0.058;0.047;0.046;0.064;0.059;0.044;0.049;0.041;0.033;0.032 \\
\cmidrule{2-5}          & \multirow{2}[4]{*}{lessrank} & \multirow{2}[4]{*}{ptb} & Ours  & \multicolumn{1}{c}{\textbf{0.056;0.044;0.048;0.059;0.050;0.054;0.061;0.064;0.064;0.068;0.052;0.051;0.057;0.049;0.059;0.055;0.031;0.031;0.025;0.025}} \\
\cmidrule{4-5}          &       &       & Shapley & 0.040;0.034;0.073;0.062;0.042;0.053;0.068;0.047;0.068;0.042;0.064;0.045;0.062;0.054;0.051;0.066;0.027;0.045;0.027;0.030 \\
\cmidrule{2-5}          & \multicolumn{1}{c}{\multirow{2}[4]{*}{random}} & \multicolumn{1}{c}{\multirow{2}[4]{*}{ptb}} & Ours  & \multicolumn{1}{c}{\textbf{0.059;0.050;0.050;0.056;0.050;0.054;0.061;0.060;0.063;0.061;0.060;0.057;0.058;0.049;0.061;0.056;0.023;0.024;0.024;0.023}} \\
\cmidrule{4-5}          &       &       & Shapley & 0.056;0.058;0.057;0.056;0.054;0.058;0.060;0.060;0.057;0.056;0.052;0.057;0.055;0.058;0.055;0.055;0.025;0.022;0.024;0.023 \\
    \bottomrule
    \end{tabular}}
\end{table*}
\begin{figure}[htbp]
	\begin{minipage}[c]{0.5\textwidth}
	\begin{center}
		\includegraphics[width=2.5in, keepaspectratio]{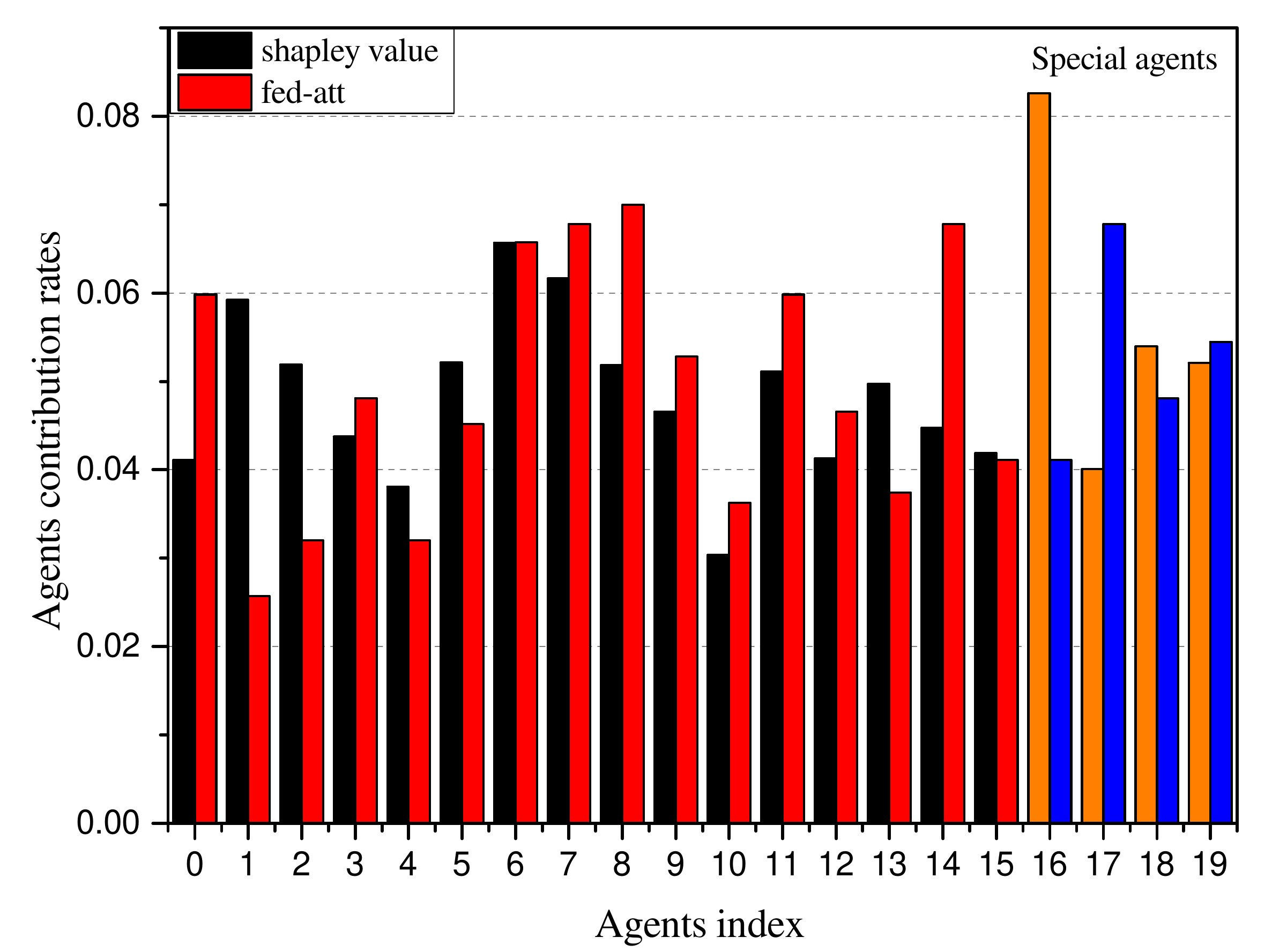}
	\end{center}
	\caption{Experimental results of normal measurement.}
	\label{ptb-normal}
	\end{minipage}
	\begin{minipage}[c]{0.5\textwidth}
		\begin{center}
			\includegraphics[width=2.5in, keepaspectratio]{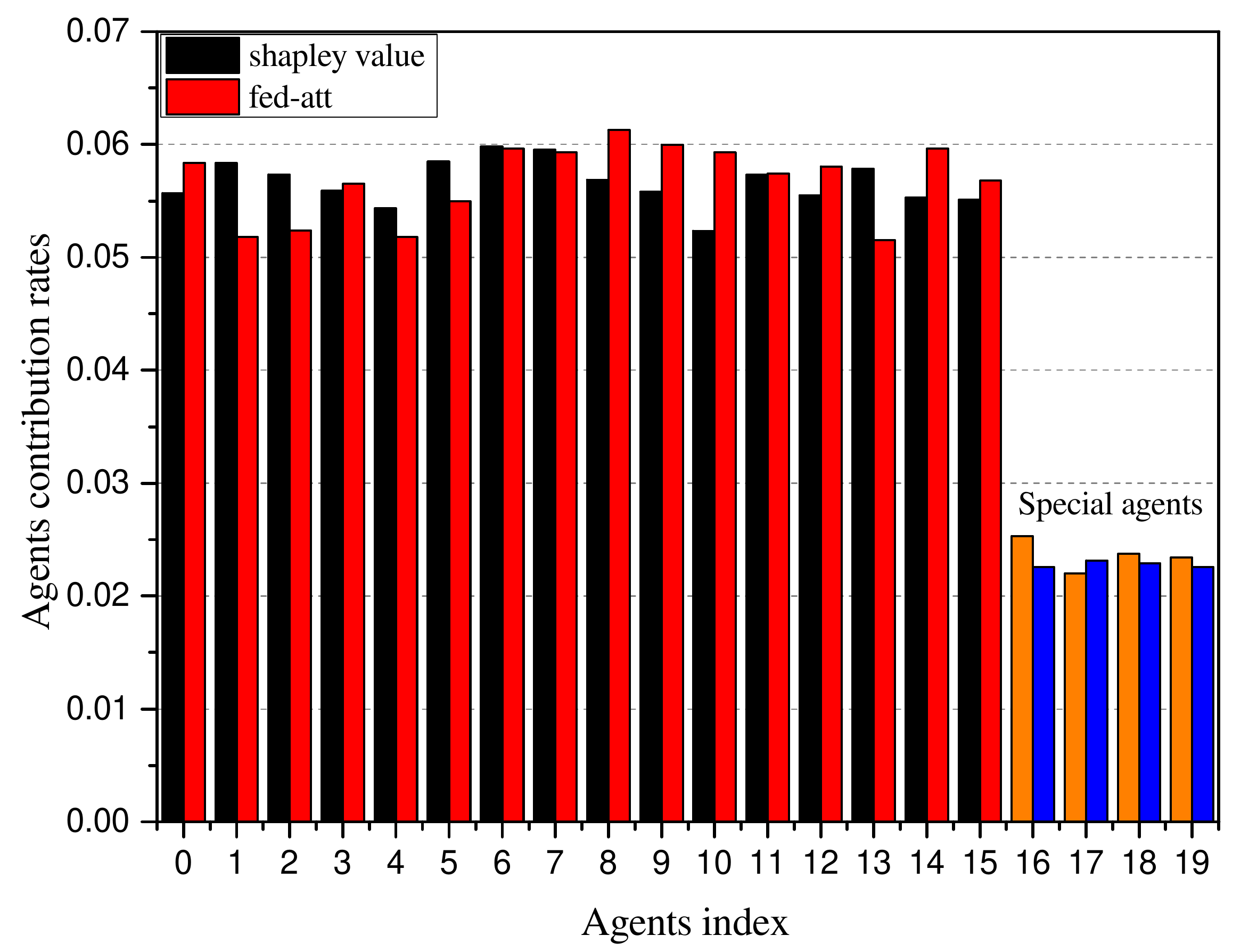}
		\end{center}
		\caption{Experimental results of randomize word sequence case.}
		\label{ptb-random}
	\end{minipage}
\end{figure}

\begin{figure}[htbp]
	\begin{minipage}[c]{0.5\textwidth}
	\begin{center}
		\includegraphics[width=2.5in, keepaspectratio]{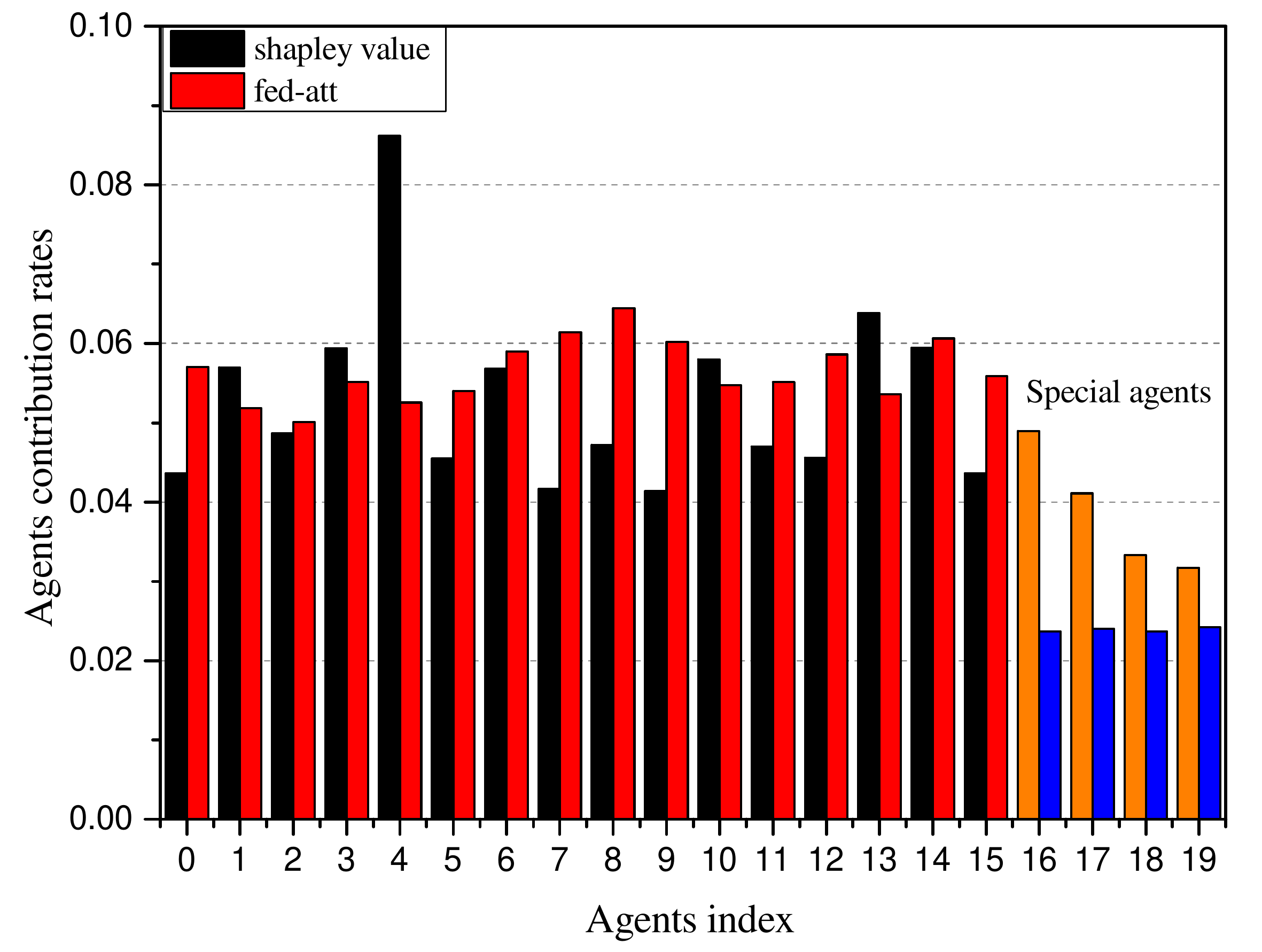}
	\end{center}
	\caption{Experimental results of reduce data case.}
	\label{ptb-less}
	\end{minipage}
	\begin{minipage}[c]{0.5\textwidth}
	\begin{center}
		\includegraphics[width=2.5in, keepaspectratio]{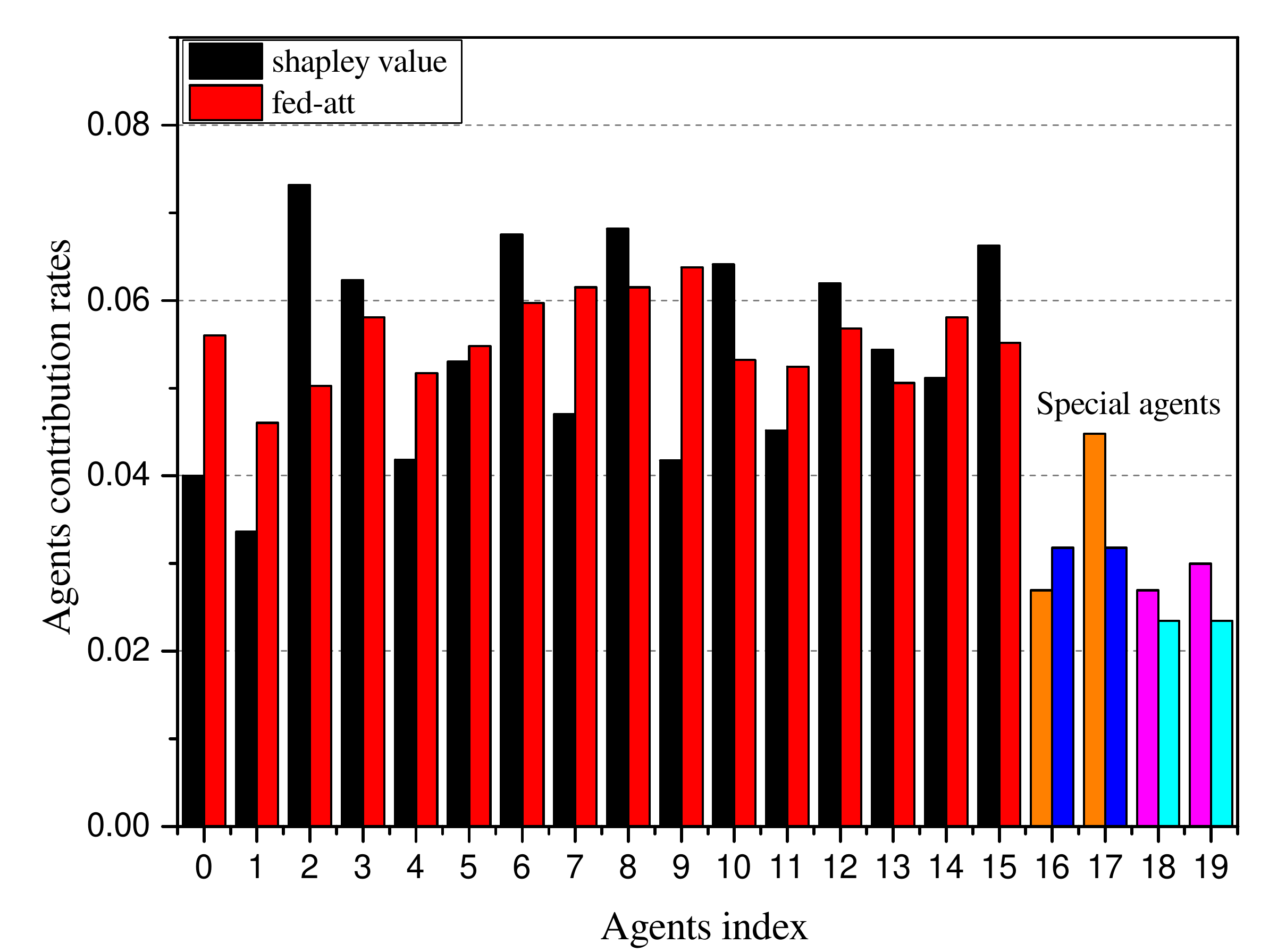}
	\end{center}
	\caption{Experimental results of reduce data different levels.}
	\label{ptb-lessrank}
	\end{minipage}
\end{figure}

We randomly divided the data in the dataset and distributed them evenly to all agents. In the language model, the amount of data will be more important, so that the model can be expressed to a meaningful degree. In this experiment we imagined three kinds of different scenarios: reduce the amount of data by 30\% and 70\%, randomly generate the word sequence. Comparing the measurement results after special processing with the results when unprocessed data, it is concluded that the sensitivity of variables such as data quantity and data quality is evaluated.\\
\textbf{-- Experimental results of normal measurement}

We did not do any special treatment to any agents and randomly divided the dataset into each agent. The result is shown in Fig. \ref{ptb-normal}. It can be seen that the deviation of each agent is not large, only individual agents are too high or too low. And the measurement result is similar to Shapley Value.\\
\textbf{-- Experimental results of random word sequence case}

In this experiment, we modified the datasets of the last four agents into randomize word sequences. These should be regarded as dirty data by the model, so as to get a smaller return. The results are shown in Fig. \ref{ptb-random}. Both the method proposed in this paper and the Shapley Value method can identify these bad agents and give them a small contribution.\\
\textbf{-- Experimental results of the reducing data case}

To demonstrate the sensitivity of our measurement method to the amount of data, we performed the experiment. In this experiment, we reduced the data amount of the last 4 agents by 70\%, and the data amount of other agents remained unchanged. As seen in Fig. \ref{ptb-less}, the contributions of the specially treated agents are significantly reduced. But it is relatively not obvious in the measurement results of Shapley Value. 

In order to reflect the relative relationship between the amount of data, we processed the data of the last 4 agents: agents with index 16 and 17 reduced the amount of data by 30\%, agents with index 18 and 19 reduced the amount of data by 70\%. It can be seen in Fig. \ref{ptb-lessrank} that the method proposed in this paper can reflect the reduced amount of data, while Shapley Value cannot show it well. Note that the method proposed in this paper can better reflect the difference of smaller data volume.

\section{Federated learning software with FedCM}
In addition to proposing an effective federated learning approach, we also developed a software ``FedMedical'' based on FedCM. FedMedical is written in Python and can be deployed on Linux, using the PaddleFL framework as the back-end of the neural network to perform parameter updates and aggregation. FedMedical also reserves the contribution measurement interface for implementation based on the FedCM method and the deployment of some incentive mechanisms. We will continue to improve these in the future.

\begin{figure}[htbp]
	\begin{center}
		\includegraphics[width=0.24\textwidth, keepaspectratio]{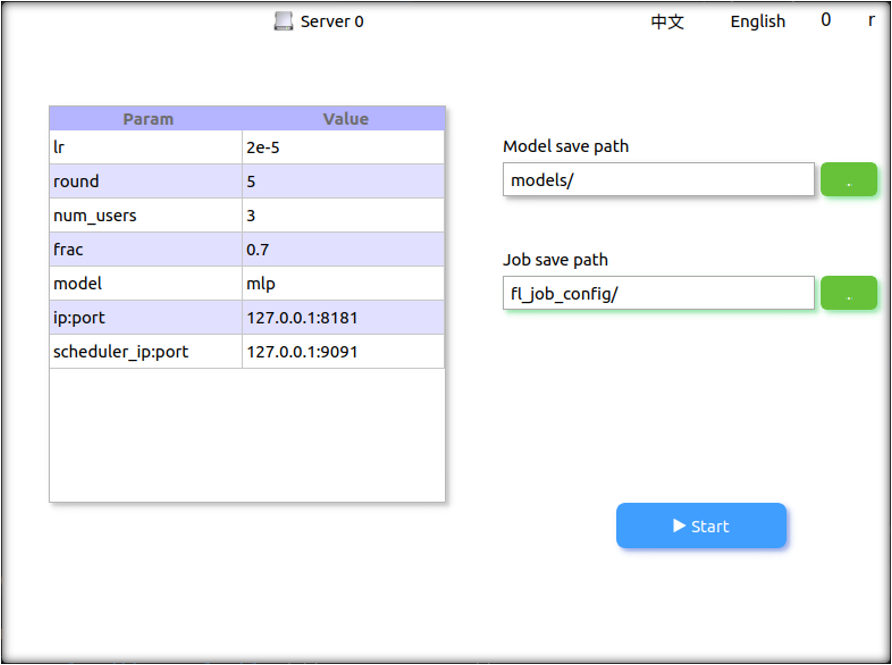}
		\includegraphics[width=0.24\textwidth, keepaspectratio]{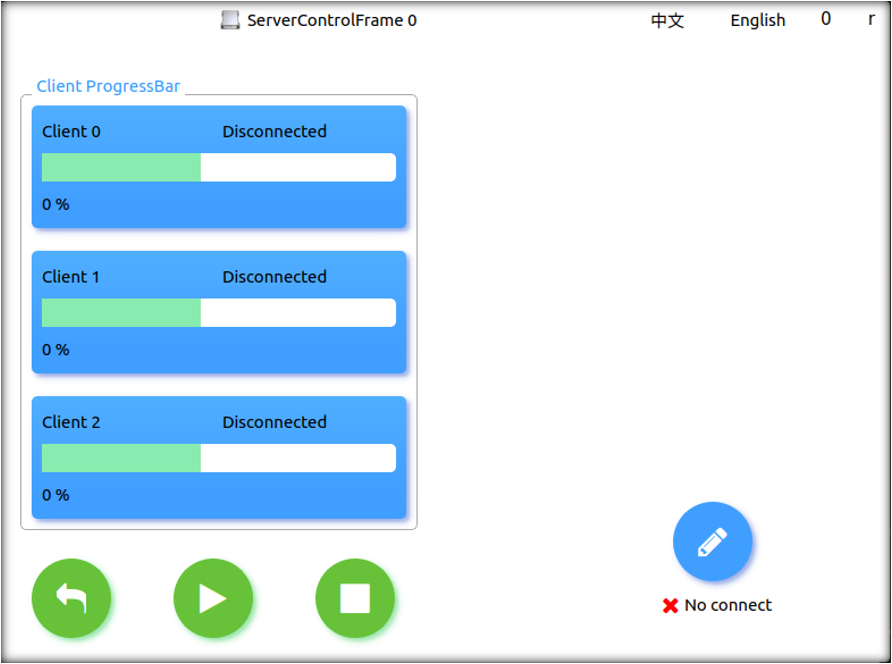}
	\end{center}
	\caption{Interfaces of FedMedical.}
	\label{fedmed}
\end{figure}

\begin{figure}[htbp]	
	\begin{center}
		\includegraphics[width=0.24\textwidth, keepaspectratio]{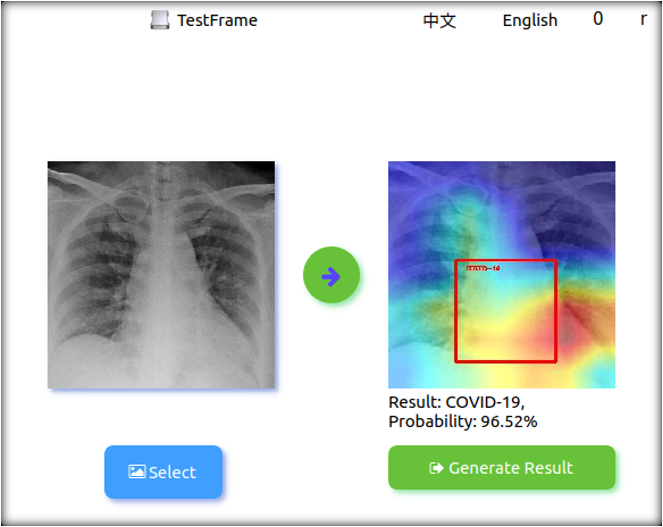}
		\includegraphics[width=0.24\textwidth, keepaspectratio]{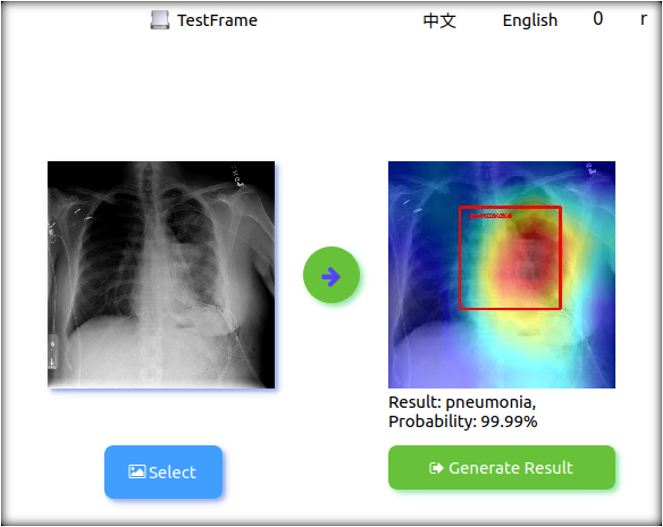}
	\end{center}
	\caption{Detection results of FedMedical.}
	\label{fedmed-2}
\end{figure}

At present, we have applied it in the medical field, such as COVID-19 detection based on medical images. As shown in the figures above, they are interfaces of the software FedMedical. FedMedical can realize COVID-19 detection based on medical images under the premise of privacy protection based on the federated learning framework. This is of great significance for shared medicine, because FedMedical can use medical data from different places to train powerful AI models for disease detection. And in this process, FedMedical completely saves the patient's data locally. If hospitals around the world can apply FedMedical, the accuracy of our disease detection model will be greatly improved.

\section{CONCLUSION}
Federated learning is a machine learning framework for protecting distributed data privacy and has participated in commercial activities. However, there is a lack of a sufficiently fair contribution measurement mechanism to distribute each agent's reward. To address this issue, we propose a real-time contribution measurement method FedCM. FedCM defines the impact of each agent, comprehensively considers the current round and the previous round to obtain each agent's contribution rate based on attention aggregation. Moreover, it is capable of performing in real-time which is important in the real federated learning system. We conduct sufficient experiments to verify the effects of FedCM. Moreover, the federated learning software Fedmedical is developed and open sourced for disease detection based on medical images. 

For future work, we expect some more advanced algorithms can be found that can more accurately measure the contribution to malicious agents, and at the same time prevent the contribution confusion caused by more than half of malicious agents. It also should consider that the negative contribution of agents measurement and the comprehensiveness of the measurement. Preliminary classification (i.e., positive and negative) is carried out at the end to avoid attacks under the federated learning framework mechanism\cite{bagdasaryan2018backdoor}; at the same time, the concept of game theory is introduced, such as the establishment of a PVCG mechanism on the supply side\cite{cong2020optimal}. Improvements in these directions will promote the implementation and application of the federated learning incentive mechanism.

%\subsubsection*{Acknowledgments}
%Use unnumbered third level headings for the acknowledgments. All
%acknowledgments, including those to funding agencies, go at the end of the paper.
\bibliography{ref}

\begin{thebibliography}{10}
\providecommand{\url}[1]{#1}
\csname url@samestyle\endcsname
\providecommand{\newblock}{\relax}
\providecommand{\bibinfo}[2]{#2}
\providecommand{\BIBentrySTDinterwordspacing}{\spaceskip=0pt\relax}
\providecommand{\BIBentryALTinterwordstretchfactor}{4}
\providecommand{\BIBentryALTinterwordspacing}{\spaceskip=\fontdimen2\font plus
\BIBentryALTinterwordstretchfactor\fontdimen3\font minus
  \fontdimen4\font\relax}
\providecommand{\BIBforeignlanguage}[2]{{%
\expandafter\ifx\csname l@#1\endcsname\relax
\typeout{** WARNING: IEEEtran.bst: No hyphenation pattern has been}%
\typeout{** loaded for the language `#1'. Using the pattern for}%
\typeout{** the default language instead.}%
\else
\language=\csname l@#1\endcsname
\fi
#2}}
\providecommand{\BIBdecl}{\relax}
\BIBdecl

\bibitem{ion2016metamaterial}
A.~Ion, J.~Frohnhofen, L.~Wall, R.~Kovacs, M.~Alistar, J.~Lindsay, P.~Lopes,
  H.-T. Chen, and P.~Baudisch, ``Metamaterial mechanisms,'' in
  \emph{Proceedings of the 29th annual symposium on user interface software and
  technology}, 2016, pp. 529--539.

\bibitem{wang2019measure}
G.~Wang, C.~X. Dang, and Z.~Zhou, ``Measure contribution of participants in
  federated learning,'' in \emph{2019 IEEE International Conference on Big Data
  (Big Data)}.\hskip 1em plus 0.5em minus 0.4em\relax IEEE, 2019, pp.
  2597--2604.

\bibitem{zhan2020learning}
Y.~Zhan, P.~Li, Z.~Qu, D.~Zeng, and S.~Guo, ``A learning-based incentive
  mechanism for federated learning,'' \emph{IEEE Internet of Things Journal},
  2020.

\bibitem{hochreiter1997long}
S.~Hochreiter and J.~Schmidhuber, ``Long short-term memory,'' \emph{Neural
  computation}, vol.~9, no.~8, pp. 1735--1780, 1997.

\bibitem{brendan2016communication}
H.~Brendan~McMahan, E.~Moore, D.~Ramage, S.~Hampson, and B.~Ag{\"u}era~y Arcas,
  ``Communication-efficient learning of deep networks from decentralized
  data,'' \emph{arXiv preprint arXiv:1602.05629}, 2016.

\bibitem{felbab2019optimization}
V.~Felbab, P.~Kiss, and T.~Horv{\'a}th, ``Optimization in federated learning.''
  in \emph{ITAT}, 2019, pp. 58--65.

\bibitem{ji2019learning}
S.~Ji, S.~Pan, G.~Long, X.~Li, J.~Jiang, and Z.~Huang, ``Learning private
  neural language modeling with attentive aggregation,'' in \emph{2019
  International Joint Conference on Neural Networks (IJCNN)}.\hskip 1em plus
  0.5em minus 0.4em\relax IEEE, 2019, pp. 1--8.

\bibitem{10.1145/3298981}
\BIBentryALTinterwordspacing
Q.~Yang, Y.~Liu, T.~Chen, and Y.~Tong, ``Federated machine learning: Concept
  and applications,'' \emph{ACM Trans. Intell. Syst. Technol.}, vol.~10, no.~2,
  Jan. 2019. [Online]. Available: \url{https://doi.org/10.1145/3298981}
\BIBentrySTDinterwordspacing

\bibitem{mcmahan2017communication}
B.~McMahan, E.~Moore, D.~Ramage, S.~Hampson, and B.~A. y~Arcas,
  ``Communication-efficient learning of deep networks from decentralized
  data,'' in \emph{Artificial Intelligence and Statistics}.\hskip 1em plus
  0.5em minus 0.4em\relax PMLR, 2017, pp. 1273--1282.

\bibitem{hardy2017private}
S.~Hardy, W.~Henecka, H.~Ivey-Law, R.~Nock, G.~Patrini, G.~Smith, and
  B.~Thorne, ``Private federated learning on vertically partitioned data via
  entity resolution and additively homomorphic encryption,'' \emph{arXiv
  preprint arXiv:1711.10677}, 2017.

\bibitem{chen2020fedhealth}
Y.~Chen, X.~Qin, J.~Wang, C.~Yu, and W.~Gao, ``Fedhealth: A federated transfer
  learning framework for wearable healthcare,'' \emph{IEEE Intelligent
  Systems}, 2020.

\bibitem{aono2017privacy}
Y.~Aono, T.~Hayashi, L.~Wang, S.~Moriai \emph{et~al.}, ``Privacy-preserving
  deep learning via additively homomorphic encryption,'' \emph{IEEE
  Transactions on Information Forensics and Security}, vol.~13, no.~5, pp.
  1333--1345, 2017.

\bibitem{lu2019blockchain}
Y.~Lu, X.~Huang, Y.~Dai, S.~Maharjan, and Y.~Zhang, ``Blockchain and federated
  learning for privacy-preserved data sharing in industrial iot,'' \emph{IEEE
  Transactions on Industrial Informatics}, vol.~16, no.~6, pp. 4177--4186,
  2019.

\bibitem{kim2019blockchained}
H.~Kim, J.~Park, M.~Bennis, and S.-L. Kim, ``Blockchained on-device federated
  learning,'' \emph{IEEE Communications Letters}, vol.~24, no.~6, pp.
  1279--1283, 2019.

\bibitem{konevcny2016federated}
J.~Kone{\v{c}}n{\`y}, H.~B. McMahan, F.~X. Yu, P.~Richt{\'a}rik, A.~T. Suresh,
  and D.~Bacon, ``Federated learning: Strategies for improving communication
  efficiency,'' \emph{arXiv preprint arXiv:1610.05492}, 2016.

\bibitem{ren2019federated}
J.~Ren, H.~Wang, T.~Hou, S.~Zheng, and C.~Tang, ``Federated learning-based
  computation offloading optimization in edge computing-supported internet of
  things,'' \emph{IEEE Access}, vol.~7, pp. 69\,194--69\,201, 2019.

\bibitem{bao2019flchain}
X.~Bao, C.~Su, Y.~Xiong, W.~Huang, and Y.~Hu, ``Flchain: A blockchain for
  auditable federated learning with trust and incentive,'' in \emph{2019 5th
  International Conference on Big Data Computing and Communications
  (BIGCOM)}.\hskip 1em plus 0.5em minus 0.4em\relax IEEE, 2019, pp. 151--159.

\bibitem{geyer2017differentially}
R.~C. Geyer, T.~Klein, and M.~Nabi, ``Differentially private federated
  learning: A client level perspective,'' \emph{arXiv preprint
  arXiv:1712.07557}, 2017.

\bibitem{xu2019elfish}
Z.~Xu, Z.~Yang, J.~Xiong, J.~Yang, and X.~Chen, ``Elfish: Resource-aware
  federated learning on heterogeneous edge devices,'' \emph{arXiv preprint
  arXiv:1912.01684}, 2019.

\bibitem{kang2019incentive}
J.~Kang, Z.~Xiong, D.~Niyato, S.~Xie, and J.~Zhang, ``Incentive mechanism for
  reliable federated learning: A joint optimization approach to combining
  reputation and contract theory,'' \emph{IEEE Internet of Things Journal},
  vol.~6, no.~6, pp. 10\,700--10\,714, 2019.

\bibitem{bagdasaryan2018backdoor}
E.~Bagdasaryan, A.~Veit, Y.~Hua, D.~Estrin, and V.~Shmatikov, ``How to backdoor
  federated learning. corr,'' \emph{arXiv preprint arXiv:1807.00459}, 2018.

\bibitem{cong2020optimal}
M.~Cong, X.~Weng, H.~Yu, J.~Qu, and S.~M. Yiu, ``Optimal procurement auction
  for cooperative production of virtual products: Vickrey-clarke-groves meet
  cremer-mclean,'' 2020.

\end{thebibliography}
\bibliographystyle{IEEEtran}
\end{document}